\documentclass[runningheads]{llncs}
\usepackage{graphicx}
\usepackage{amsmath}

\usepackage{tabularray}
\usepackage{array}
\newcommand{\repeatthanks}{\textsuperscript{\thefootnote}}
\usepackage[
colorlinks=true,
urlcolor=blue,
linkcolor=green
]{hyperref}

\begin{document}
\title{Multimodal Analysis of White Blood Cell Differentiation in Acute Myeloid Leukemia Patients using a $\beta$-Variational Autoencoder}
\titlerunning{Multimodal Analysis of WBC}

\author{Gizem Mert\inst{1,2}\thanks{equal contribution}
\and Ario Sadafi\inst{1,3}\repeatthanks
\and Raheleh Salehi\inst{1,4}
\and \\Nassir Navab\inst{3,5}
\and Carsten Marr\inst{1}\thanks{carsten.marr@helmholtz-munich.de}
}
%index{Mert, Gizem}
%index{Sadafi, Ario}
%index{Salehi, Raheleh}
%index{Navab, Nassir}
%index{Marr, Carsten}

\authorrunning{G. Mert and A. Sadafi et al.}

\institute{Institute of AI for Health, Helmholtz Zentrum München – German Research Center for Environmental Health, Neuherberg, Germany
\and 
Department of Information Engineering and Computer Science, University of Trento, Trento, Italy
\and
Computer Aided Medical Procedures (CAMP), Technical University of Munich, Munich, Germany
\and
Institute for Chemical Epigenetics Munich (ICEM), Department of Chemistry, Ludwig-Maximilians University Munich, Munich, Germany
\and
Computer Aided Medical Procedures, Johns Hopkins University, Baltimore, USA
}
\maketitle       
\begin{abstract}
    Biomedical imaging and RNA sequencing with single-cell resolution improves our understanding of white blood cell diseases like leukemia. By combining morphological and transcriptomic data, we can gain insights into cellular functions and trajectoriess involved in blood cell differentiation. However, existing methodologies struggle with integrating morphological and transcriptomic data, leaving a significant research gap in comprehensively understanding the dynamics of cell differentiation. Here, we introduce an unsupervised method that explores and reconstructs these two modalities and uncovers the relationship between different subtypes of white blood cells from human peripheral blood smears in terms of morphology and their corresponding transcriptome. Our method is based on a beta-variational autoencoder ($\beta$-VAE) with a customized loss function, incorporating a R-CNN architecture to distinguish single-cell from background and to minimize any interference from artifacts. This implementation of $\beta$-VAE shows good reconstruction capability along with continuous latent embeddings, while maintaining clear differentiation between single-cell classes. Our novel approach is especially helpful to uncover the correlation of two latent features in complex biological processes such as formation of granules in the cell (granulopoiesis) with gene expression patterns. It thus provides a unique tool to improve the understanding of white blood cell maturation for biomedicine and diagnostics.

\end{abstract}

\section{Introduction}
Hematopoietic stem cells are versatile: They can develop into all types of mature blood cells. These emerge from two main branches: the myeloid branch includes the common myeloid progenitor and several subsequent cell types. 
The lymphoid branch includes the common lymphoid progenitor, which gives rise to various types of lymphocytes, including T cells, B cells, and natural killer cells.
Leukemia is a type of cancer that affects blood cells, typically starting in the bone marrow and leading to the proliferation of immature blood cells \cite{redecke2013hematopoietic}.
Acute myeloid leukemia (AML) is a group of leukemias that arise from the myeloid branch \cite{pelcovits2020acute}.

Hematopoietic cells are traditionally identified by their morphology through microscopic examination of blood smears, a method that relies on the expertise of medical professionals to detect abnormalities indicative of leukemia. Concurrently, advancements in machine learning have enabled the automatic classification of blood cell types using large datasets of single-cell images. For example, Matek et al. \cite{matek2019human} developed a method for human-level classification of blast cells in AML patients, demonstrating high accuracy in analyzing bone marrow smears \cite{matek2021highly}. 
% These automated approaches can enhance leukemia diagnosis by providing consistent and objective analysis, thus supporting medical professionals in making timely and accurate diagnoses. 

From a genetic perspective, driver genes play a crucial role in identifying and distinguishing between various subtypes of leukemia. Mutations in genes and chromosomal fusions or displacements serve as drivers for subtype identification. Leukemia subtypes are often named based on these genetic alterations, such as AML with NPM1 mutation or AML with RUNX1::RUNX1T1 fusion \cite{khoury20225th}. These genetic signatures provide important insights into the underlying mechanisms and characteristics of different leukemia subtypes as well as subtle morphological differences in the cells \cite{hehr2023explainable,sadafi2023pixel}.

High-throughput gene expression (transcriptomic) analyses are widely utilized across various domains of biomedicine, ranging from fundamental research to molecular diagnosis \cite{gottschlich2023single,sikkema2023integrated,park2020prenatal,knoll2024life}.  RNA sequencing (RNA-seq) is a sequencing technique used to analyze the quantity and sequences of RNA in a sample. 
RNA-seq analysis, while powerful and capable of revealing numerous exciting discoveries, diverges from the typical analyses familiar to bench scientists in that it presents as a vast dataset that requires extensive analysis for interpretation \cite{koch2018beginner}. 
Consequently, several different atlases have been constructed using the abundant data made available through RNA-seq analyses \cite{tomczak2015review,suo2022mapping,park2020cell}. 

Autoencoders serve as potent tools for generating latent representations of data. They encode crucial information from input data, enabling the decoder to reconstruct the same data using lower-dimensional information that has passed through the bottleneck. While traditional autoencoders encode data into a single point, variational autoencoders (VAEs) encode input into a distribution in the latent space, allowing the formation of continuous manifolds from which one can sample. $\beta$-VAEs \cite{higgins2016beta} are particularly powerful for learning disentangled latent representations. In a disentangled representation, a variable in the latent space is sensitive only to one generative factor and remains invariant to other factors. Such representations are valuable for interpretability and generalize well across various tasks.

\begin{figure}[t]
\centering
\includegraphics[width=\textwidth,page=1,trim=0cm 0cm 0cm 0cm,clip]{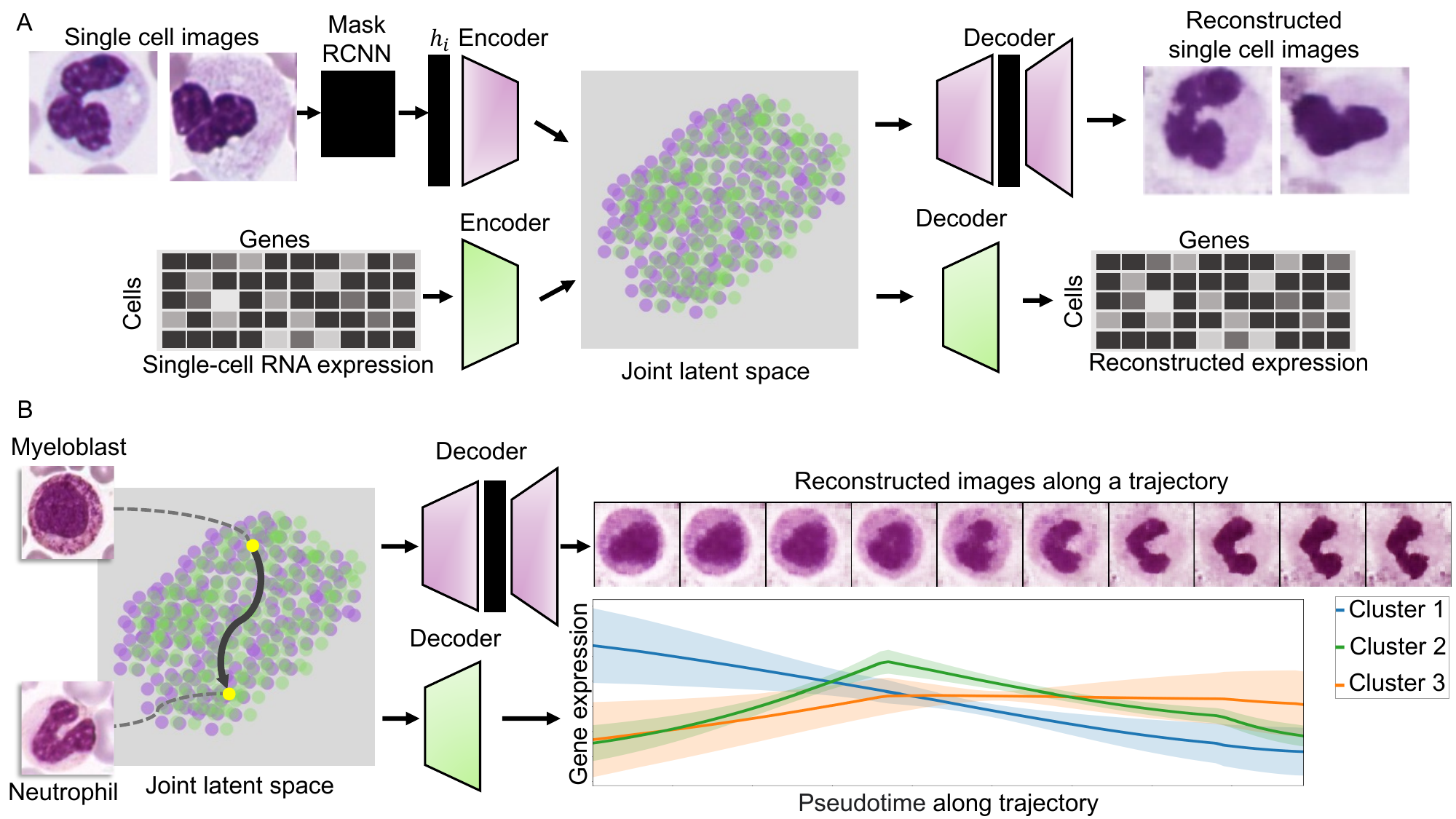}
\caption{Our method is able to jointly embed and reconstruct single-cell image data and transcriptomic information. A) single-cell images are analyzed using a mask R-CNN for feature extraction and are embedded into a joint latent space with encoded single-cell RNA expressions. B) Interpolations defined on the latent space can be decoded into single-cell images and gene expression variation along the trajectory. Expression variations are clustered showing genes categorized based on different transcriptomic kinetics.}
\label{fig:over}
\end{figure}

In this paper, we propose a novel approach based on $\beta$-VAE to jointly embed morphological and transcriptomic data of hematopoietic cells. This marks the first instance where researchers can simulate the transition from progenitor cells, like myelocytes, into mature cells like neutrophils and monocytes (See Fig \ref{fig:over}).
The process of transforming a stem cell into a mature neutrophil involves intermediate steps (promyelocyte, myelocyte, metamyelocyte, and band and segmented neutrophils), occurring in different organs like bone marrow, lung, liver, and peripheral blood. Our analysis can unveil distinct correlations between morphology and variations in gene expressions from transcriptomic data, aiding to understand the intricate details of the white blood cell production process.

\section{Methods}
Our method relies on a $\beta$-VAE framework. The encoder takes in the features and maps them into a 50-dimensional latent space, while a two-staged decoder attempts to reconstruct both the features and the corresponding single-cell image (Fig \ref{fig:over}A). A mask generated by the mask R-CNN \cite{he2017mask} is employed for the reconstruction and embedding process to mitigate the influence of artifacts, like red blood cells surrounding the white blood cell in an image.
A second $\beta$-VAE is employed to embed single-cell RNA sequencing (scRNA-seq) data into the same latent space. To address potential batch effects or issues stemming from sparse and non-normalized data, we utilize the Tabula Sapiens atlas \cite{tabula2022tabula}. Tabula Sapiens is a molecular reference atlas of the human body consisting of over 400 cell types. Various issues known in scRNA-seq data such as batch effect or low gene counts have been addressed when the atlas has been generated. The embeddings from the two autoencoders are aligned using maximum mean discrepancy \cite{salehi2022unsupervised}. With two reference classes established, an interpolation is defined between the two points in the latent space using Gaussian Process regression \cite{rasmussen2006gaussian}. Subsequently, the two decoders generate corresponding morphological changes as well as trajectories in gene expressions, allowing us to observe the migration of the cells over time in the latent space, which corresponds to blood cell differentiation.

Our goal is to create a joint latent manifold $\mathcal{P}$ derived from both cell images and scRNA-seq data. To achieve this, we extract features $h_i$ using RoIAlign and obtain corresponding binary masks $m_i$ for each single-cell image $I$ through the Mask R-CNN model. A latent distribution $\mathcal{N}(\mu^{\mathrm{img}},\sigma^{\mathrm{img}^2}) \in \mathcal{P}$ is derived by the image encoder for each image. This distribution is obtained as $\mu, \sigma = \mathrm{f}_{\mathrm{img\_enc}}(h_i; \theta)$, where $\theta$ represents the parameters of the encoder. The model is trained by minimizing the loss function

\begin{equation}
\begin{split}
    \mathcal{L}_{\mathrm{img}}(\theta, \eta, \tau) = \alpha \mathrm{MSE}(\hat{h_i}, h_i) + \gamma \mathrm{MSE}(m_i\hat{I_i}, m_iI_i) + \\ \beta \mathcal{L}_\mathrm{KL}(\theta) + \delta \mathcal{L}_\mathrm{align} (\theta) ,
    \label{eq:1}
\end{split}
\end{equation}
where $\hat{h_i} = \mathrm{f_{feat\_dec}}(z; \eta)$ and $\hat{I_i} = \mathrm{f_{img\_dec}}(h_i; \tau)$ are the feature and image reconstructions obtained by feature image decoders respectively. $\alpha$, $\gamma$, $\beta$, and $\delta$ are coefficients regulating the impact of each term on the total loss, and $z = \mu + \sigma \odot \epsilon$ is a point obtained through reparameterization trick \cite{kingma2015variational} for decoding. The loss for image reconstruction is focused only on the cell region using the mask $m_i$ and $\theta, \eta, \tau$ are learnable parameters optimized during the training.  $\mathrm{MSE}(.,.)$ is the mean squared error function.
\begin{equation}
    \mathcal{L}_\mathrm{KL} (\theta) = \frac{1}{2} \sum (\exp(\sigma^{{\mathrm{img}}^2}) + \mu^{\mathrm{img}^{2}} -1 - \mathrm{log}(\sigma^{2})
    \label{eq:kl}
\end{equation}
is the Kullback-Leibler divergence defined between the obtained distribution and a reference normal distribution. This term is regulated with a coefficient $\beta$ in equation \ref{eq:1} for better disentanglement of the latent dimensions.

We have selected some of the cells relevant to our study and use their embeddings in Tabula Sapiens (TS) atlas as reference to form the latent space.
If a point of the atlas is $z^{\mathrm{TS}}$ and $X_i$ is the gene expression matrix of a given cell, a latent embedding $\mu^\mathrm{rna}, \sigma^\mathrm{rna} =\mathrm{f}_{\mathrm{rna\_enc}}(X_i; \rho)$ such that $\mathcal{N}(\mu^{\mathrm{rna}},\sigma^{\mathrm{rna}^2}) \in \mathcal{P}$ is obtained by the RNA encoder by optimizing the following loss

\begin{equation}
    \mathcal{L}_{rna} (\rho, \omega)= \lambda \mathrm{MSE}(X,\hat{X}) + \phi (1 - \mathrm{cosine}(z, z^{\mathrm{TS}})) + \beta \mathcal{L}_{\mathrm{KL}} (\rho)
\end{equation}
where $\hat{X} = \mathrm{f_{rna\_dec}}(z; \omega)$ and $z$ is similarly calculated using a reparametrization trick. $\mathcal{L}_{\mathrm{KL}}$ is the same as Eq. \ref{eq:kl} and cosine is the cosine similarity loss. The mean squared error $\mathrm{MSE}$ is used with a coefficient $\lambda$  for the loss between reconstructed $\hat{X}$ and original \textit{X} gene expressions. 

Finally, to align the two modalities we define a maximum mean discrepancy loss $\mathcal{L}_\mathrm{align}$ as
\begin{equation}
    \mathcal{L}_\mathrm{align} (\theta) = \sum_{j=1}^{J} \mathrm{MSE}(\mu_{j}^{\mathrm{img}},\mu_{j}^{\mathrm{rna}}) + \mathcal{L}_\mathrm{KL} (s_{j}^\mathrm{img}, s_{j}^\mathrm{rna}) 
\end{equation}
where $\mu_{j}$ and $s_{j}$ are mean and softmax of the covariance matrix of the embeddings of every class $j$ in RNA and image modalities, and $D_\mathrm{KL}$ is the Kullback-Leibler divergence.

With this setting, after optimizing all learnable parameters, we obtain a latent manifold $\mathcal{P}$. This manifold enables the study of the correlation between transcriptome and morphology of single-cells. To model cell differentiation from class A to class B in the latent space, an interpolation between two points is performed. We utilize Gaussian Process regression \cite{rasmussen2006gaussian} to calculate this interpolation.
% Having $X=[x_1,...,x_n]$ representing the observed data from $\mathcal{P}$, the regression function is modeled by a multivariate Gaussian $P(f | X) = \mathcal{N}(f|\mu, K)$ where $K$ is the RBF kernel and $f = [f(x_1),...f(x_n)]$ is the function values.
The joint probability distribution of the Gaussian Process (GP) is a multivariate Gaussian $p(y) = \mathcal{N}(m,K)$ where $m$ is the mean vector and $K$ is the covariance matrix, which is a RBF kernel function in our case. The RBF kernel $K$ calculates the similarity between two points, controlling the smoothness of the interpolation:
\begin{equation}
    K(\mathbf{x},\mathbf{x}')=\exp\left(-\frac{{\|\mathbf{x}-\mathbf{x}'\|}^2}{2\sigma^2}\right)
\end{equation}
It consists of the squared Euclidean distance between data  point $\mathbf{x}$  and $\mathbf{x}'$ and $\sigma^2$ is the variance parameter of the kernel. 

After fitting the Gaussian Process regression model, we predict the new latent vectors, obtaining a predictive distribution as 
\begin{equation}
p(\mathbf{y}|\mathbf{Y})=\mathcal{N(\mathbf{\mu_j},\mathbf{\Sigma_j})},
\end{equation}
where $\mu_j$ is the predictive mean and $\Sigma_j$ is the predictive covariance matrix. This distribution  provides a smooth path from class A to B in the latent space.

For deeper analysis of the trajectory tracks, we apply KShape clustering \cite{paparrizos2015k} to the decoded gene expression variations observed along defined trajectories (Section \ref{sec:traj}). This approach helps us identify driving genes and classify them based on their behaviors, which typically include upregulation, downregulation, or no significant change.

\section{Experiment}

\subsection{Data}
We are using two datasets to perform the experiments:

\textbf{Single-cell image data}, created in-house, contains 32,822 white blood cells from 13 different classes. Images are $288 \times 288$ pixels or $25 \times 25$ micrometers. The dataset has been annotated by medical experts and includes classes of myeloblast, promyelocyte, metamyelocyte, basophil, neutrophil banded, neutrophil segmented, monocyte, eosinophil, erythroblast, myelocyte, lymphocyte typical, lymphocyte atypical, and smudge cells. 
% Figure[?] shows the distribution of cells for this subtypes.

\textbf{The Tabula Sapiens} \cite{tabula2022tabula} is a comprehensive, multi-organ, single-cell transcriptomic atlas designed to map human gene expression at cellular level. It includes normalised and harmonised data of nearly 500,000 cells from 24 organs of 15 people, with a latent information generated using single-cell variational inference (scVI) tools \cite{lopez2018deep}. 

For our study, we focus on cells from blood circulation and lung tissue, such as myeloblasts, neutrophils, and monocytes, which align well with the cell types available in the image data. We limit the number of genes used for the study to 2,432, considering them highly variable in expression. In total, we utilize 1,862 myeloblasts, 8,820 neutrophils, and 10,973 monocytes sequences for this analysis.

\begin{figure}[t]
\centering
\includegraphics[width=\textwidth,page=3,trim=0cm 8cm 0cm 0cm,clip]{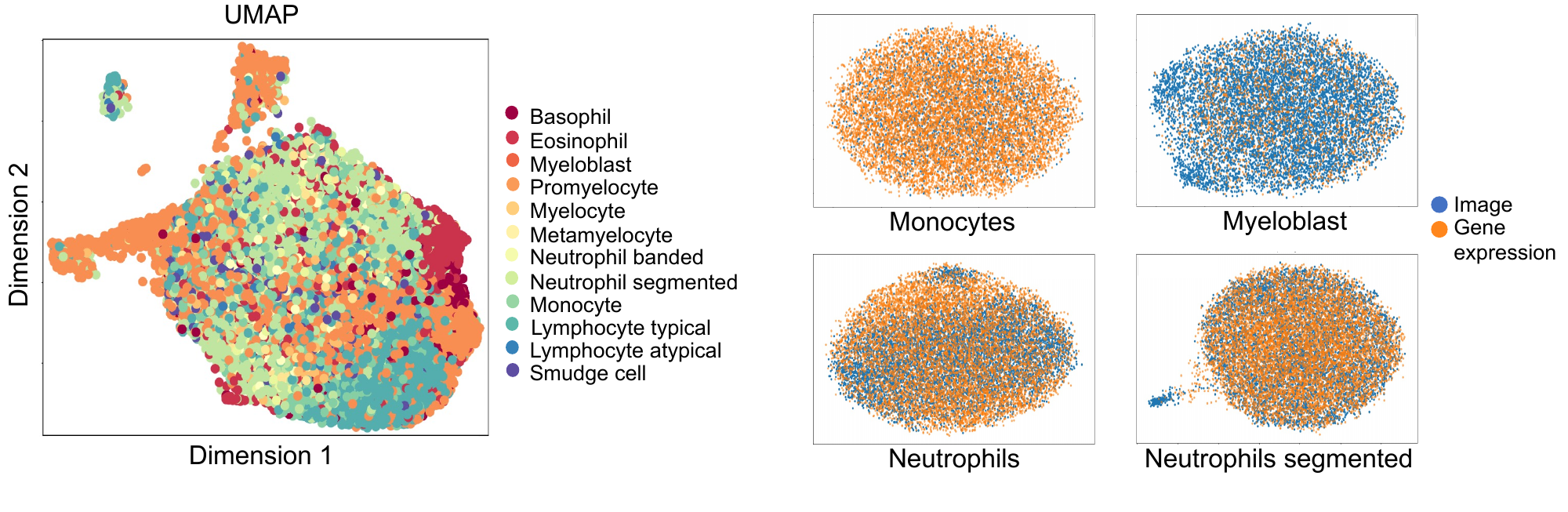}
\caption{Single-cell image data is embedded with scRNA-seq data into a continuous and morphologically separated latent space (left). Alignment of gene expressions and image data for selected classes demonstrates a close overlap enabling proper reconstruction of both modalities (right). }
\label{fig:umap}
\end{figure}

\subsection{Implementation details}
The $\beta$-VAE used for image analysis is composed of convolutional and linear layers. It accepts input features from RoIAlign with a size of $256 \times 14 \times 14$. The encoder maps these features to a 50-dimensional latent space. The decoder consists of two stages: the first stage reconstructs the features, while the second stage reconstructs the cell images. We train the model for 160 epochs using an Adam optimizer and a learning rate of 0.001.

The $\beta$-VAE for gene expression analysis employs a symmetric encoder and decoder architecture. The encoder embeds the input into a 50-dimensional latent space, similar to the image encoder. It comprises fully connected layers, with an input size of 2432, corresponding to the number of filtered genes. The network is optimized in a similar manner as the image VAE and is trained for 300 epochs.

By integrating Mask R-CNN prior to the autoencoder, we successfully mitigate the influence of surrounding artifacts on the model \cite{salehi2022unsupervised}. This strategy yields a well-distributed and continuous embedding of individual cells, as evidenced by the UMAP visualization of the latent space (Fig \ref{fig:umap}).

Furthermore, the second $\beta$-VAE accurately reconstructs gene expressions while closely adhering to the distribution of the reference atlas (Fig \ref{fig:over}). We preserve the transcriptomic latent embedding unchanged and adjust the image $\beta$-VAE accordingly. This alignment creates a joint latent space, allowing for the simultaneous decoding of gene expressions and image data.

\subsection{Trajectories in white blood cell differentiation}
\label{sec:traj}
To explore and visualize changes in morphological features and gene expression patterns simultaneously, we select initial and target points for interpolation. By generating trajectories between different cell types in the joint embedding space, we decode differentiation stages to obtain corresponding image and gene expression values along the path. This allows us to track and analyze the transitions between various cell types during white blood cell production.

\begin{figure}[t]
\centering
\includegraphics[width=\textwidth,page=2,trim=0cm 2cm 0cm 0cm,clip]{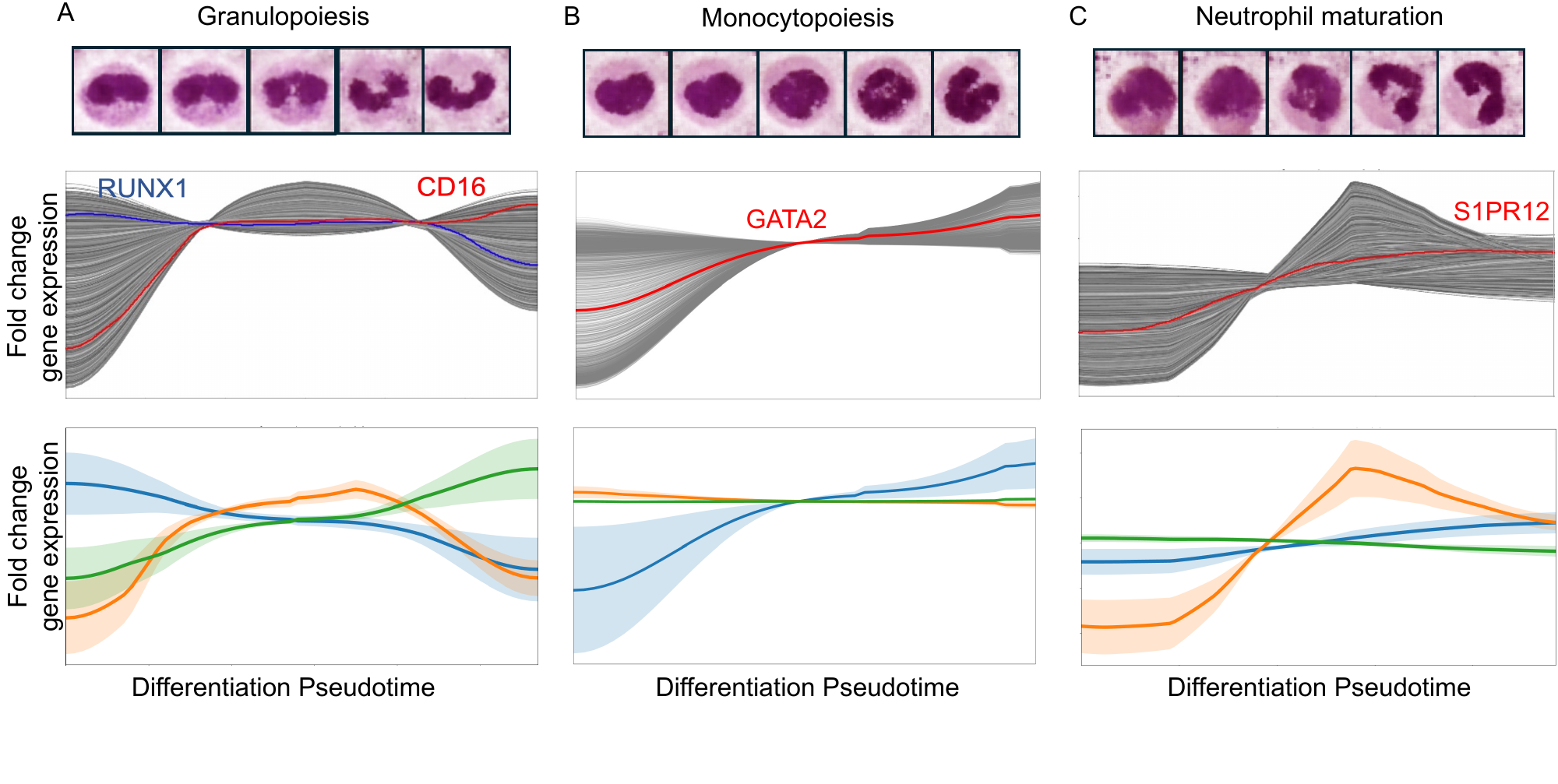}
\caption{Samples from three different trajectories defined in experiments. For every trajectory the generated images, gene expression variation, and gene expression clusters are shown. }
\label{fig:traj}
\end{figure}

\textbf{Granulopoiesis} refers to the differentiation process from myeloblast to a neutrophil cell \cite{radin2022granulopoiesis}. Neutrophils are characterized by their distinctive multi-lobed nuclei, typically containing three to five segments (Figure\ref{fig:traj}).

\textbf{Monocytopoiesis} is the differentiation from myeloid to monocyte cells. Monocytes are characterized by their distinctive bean-shaped nuclei (Fig \ref{fig:traj}).

\textbf{Neutrophil maturation} delineates the migration from banded neutrophils to segmented neutrophils, which represent the mature form of neutrophils. Both forms of neutrophils exhibit high similarity and have not been individually sequenced. Leveraging the fact that neutrophils mature in the lungs, we gain access to a greater number of banded neutrophil cells through this organ compared to the mature ones commonly found in circulation. We thus compute the interpolation from banded neutrophils to segmented neutrophils, found in lung tissues and blood circulation, respectively.

% For discovering the relations between this reconstructed single-cell images and expressions, we generate interpolations by GPR between selected points over this joint embedding. We visualise the grid of 100 generated images between two class and can see the smooth and continuous changes particularly as in Figure[?]. Also, we can see on the same figure the fold changes of each gene expressions by the mean over pseudotime(?) of the trajectory. We can cluster this genes accordingly their expression changes and discover which genes have upregulations, downregulations or showing different behaviours over the interpolation. 

\subsection{Results}
Figure\ref{fig:traj} displays three example trajectories along each of the defined white blood cell differentiation paths. The figure includes reconstructed cell images, gene expressions, and clustering results.
We compare our findings based on the clusters with genes already identified in the literature for different stages of the process. For granulopoiesis (Figure\ref{fig:traj}A), which involves the lineage from myeloblasts to neutrophils, we observe that driving genes such as RUNX1 or LEF-1 appear in the blue cluster, showing downregulation over pseudotime. Genes like CD16, MPO, and ELANE appear in the green cluster, exhibiting upregulation as expected, in line with a previous study \cite{yin2018armed}.

In the second trajectory, which represents the transition from myeloblasts to monocytes, we observe that genes like ARG1, GATA2, CREBBP, and FLI-1 are upregulated, consistent with expectations \cite{zhang2022central}. Conversely, genes like LEF-1 and RUNX1 are downregulated, aligning with known patterns \cite{kurotaki2017transcriptional}.

In the third trajectory, representing the migration from banded neutrophils in lung tissue to segmented neutrophils in blood, we observe the upregulation of the S1PR1 gene during the morphological change, consistent with the findings of Wilkins et al. \cite{wilkins2023dissecting}. With less restrictive filtration of genes based on expression values in the preprocessing step, we may capture more genes that are highly variable among the two states.

\section{Conclusion}
In this study, we propose a novel method to uncover correlations between gene expression and single-cell morphology, aiming to identify driver genes involved in white blood cell differentiation. Through our analysis, we rediscovered several genes already known in the literature to be crucial in this process. Additionally, we identified a dozen more candidate genes that require further verification by hematological experts.

However, we encountered challenges in analyzing trajectories due to limited availability of single-cell transcriptomic data. For future research, our goals include validation of candidate genes, extending our method to different tissues, and training our models on larger datasets. These endeavors will contribute to a deeper understanding of white blood cell production and its implications in disease pathology.

\bibliographystyle{splncs04}
\bibliography{Paper-007}
\end{document}